\documentclass{article}

\usepackage[preprint]{neurips_2024}
\usepackage{array}

\usepackage[utf8]{inputenc} 
\usepackage[T1]{fontenc}    
\usepackage{hyperref}       
\usepackage{url}            
\usepackage{booktabs}       
\usepackage{amsfonts}       
\usepackage{nicefrac}       
\usepackage{microtype}      
\usepackage{graphicx}
\usepackage{amsmath}
\usepackage{xspace}
\usepackage{multirow}
\usepackage{subfigure}
\usepackage{wrapfig}
\usepackage{adjustbox}
\usepackage{makecell}
\usepackage{pifont}
\usepackage{appendix}
\usepackage{bbding}
\usepackage{fontawesome} 
\usepackage{mathrsfs}
\usepackage{color, colortbl}
\usepackage[dvipsnames]{xcolor}
\definecolor{mygray}{gray}{.88}

\title{Autonomous Driving with Spiking Neural Networks}

\def\name{SAD}

\author{Rui-Jie Zhu$^{1}\thanks{Equal contribution}$, Ziqing Wang$^{2*}$, Leilani Gilpin$^{1}$, Jason K. Eshraghian$^{1}$\thanks{Corresponding author, jsn@ucsc.edu}\\ 
~\\
$^{1}$University of California, Santa Cruz, USA\\
$^{2}$Northwestern University, USA
}

\begin{document}

\maketitle

\begin{abstract}
Autonomous driving demands an integrated approach that encompasses perception, prediction, and planning, all while operating under strict energy constraints to enhance scalability and environmental sustainability. We present Spiking Autonomous Driving (\name{}), the first unified Spiking Neural Network (SNN) to address the energy challenges faced by autonomous driving systems through its event-driven and energy-efficient nature. SAD is trained end-to-end and consists of three main modules: perception, which processes inputs from multi-view cameras to construct a spatiotemporal bird's eye view; prediction, which utilizes a novel dual-pathway with spiking neurons to forecast future states; and planning, which generates safe trajectories considering predicted occupancy, traffic rules, and ride comfort. Evaluated on the nuScenes dataset, SAD achieves competitive performance in perception, prediction, and planning tasks, while drawing upon the energy efficiency of SNNs. This work highlights the potential of neuromorphic computing to be applied to energy-efficient autonomous driving, a critical step toward sustainable and safety-critical automotive technology. Our code is available at \url{https://github.com/ridgerchu/SAD}.
\end{abstract}

\section{Introduction}
Autonomous driving, often considered the `holy grail' of computer vision, integrates complex processes such as perception, prediction, and planning to achieve higher levels of vehicle automation as classified by the SAE J3016 standard~\cite{SAE2014}. Many new vehicles now feature Level 2 autonomy, with transitions to Level 3 marking notable advancements.
However, these systems must adhere to energy constraints of 50-60~W/h~\cite{powerbudget} and face increasing environmental concerns. Sudhakar \textit{et al.} highlight the need for hardware efficiency to double every 1.1 years to maintain 2050 emissions from autonomous vehicles below those of 2018 data center levels~\cite{sudhakar2022data}. 



Spiking Neural Networks (SNNs) offer a promising solution for energy-efficient intelligence by using sparse, event-driven, single-bit spiking activations for inter-neuron communication, mimicking biological neurons~\cite{roy2019towards,eshraghian2023training,li2023brain,maass1997networks}. Such workloads can be accelerated for low latency and low energy, when processed on neuromorphic hardware that utilizes asynchronous, fine-grain processing to efficiently handle spiking signals and parallel operations~\cite{davies2018loihi}. Much like in the brain, spikes are thought to encode information over time, and have shown improvements in the energy efficiency of sequence-based computer vision tasks by several orders of magnitude in a variety of workloads~\cite{azghadi2020hardware, frenkel2019morphic, ottati2023spike}.


In the past several years, SNNs have rapidly improved in performance across various tasks, including image classification~\cite{Fang_2021_ICCV,yao2024spike,zhou2022spikformer,zhu2022tcja,wang2023masked,wang2024autost,qiu2024gated,deng2024tensor,shan2023or}, object detection~\cite{su2023yolo,kim2020spikingyolo,yao2023spikev2}, semantic segmentation~\cite{kim2022segmentation}, low-level image reconstruction~\cite{qiu2023vtsnn,kamata2022vae,zhan2023esvae,qiu2023taid,cao2024spiking}, and language modeling~\cite{zhu2023spikegpt, lv2023spikebert, bal2024spikingbert}, with most of these works focused on computer vision. These advancements have brought SNN performance closer to that of Artificial Neural Networks (ANNs) in fundamental computer vision tasks. Despite this progress, SNNs have not yet proven effective in complex real-world computer vision applications that involve multiple subtasks. 


We introduce the first SNN designed for end-to-end autonomous driving, integrating perception, prediction, and planning into a single model. Achieving this milestone for SNNs involved spatiotemporal fusion of visual embeddings for enhanced perception, probabilistic future modeling for accurate prediction, and and a high performance temporal mixing spiking recurrent unit that effectively incorporates safety and comfort considerations into high-level planning decisions. 
By leveraging the event-driven and energy-efficient properties of SNNs, our model processes visual inputs, forecasts future states, and calculates the final trajectory for autonomous vehicles. 
Previously, GRUs or 3D convolutions were used in spatiotemporal visual tasks, though these operators have been entirely replaced with spiking neurons for time-mixing. 
This work marks a significant advancement in neuromorphic computing, demonstrating the potential of SNNs to handle the complex requirements of low-power autonomous driving. Our experiments show that this SNN-based system performs competitively with traditional deep learning approaches, while offering improved energy efficiency and reduced latency.

\begin{figure}
    \centering
    \includegraphics[width=\linewidth]{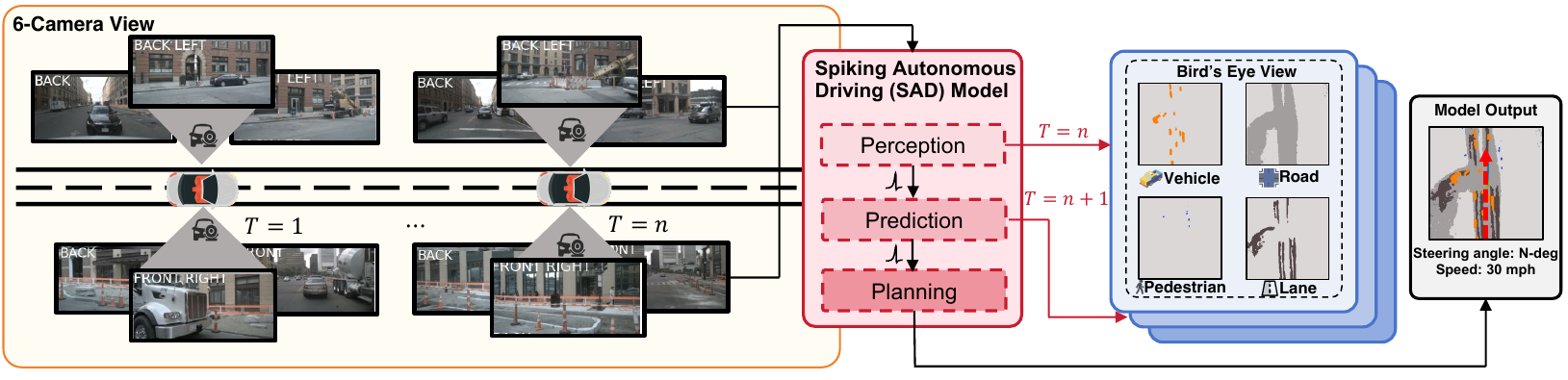}
    \caption{How SAD enables autonomous driving from vision to planning: The system processes inputs from six cameras across multiple frames. The perception module encodes feature information related to the present input frame ($T=n$), the prediction module predicts feature information of the next frame using sequential information ($T=n+1$), and the model output generates a steering and acceleration plan. This process creates a bird's eye view (BEV) and trajectory plan for navigation.}
    \label{fig:intro}
\end{figure}

\section{Related Works}\label{Sec2:relate works}

\textbf{Spiking Neural Networks in Autonomous Systems} are particularly effective for low-power, edge intelligence applications, leveraging deep learning and neuroscience principles to boost operational efficiency \cite{li2023brain,roy2019towards, henkes2022spiking, hu2021spikingresnet, schmidgall2023brain}. 
Many neuromorphic autonomous systems use SNNs as a Proportional Derivative-Integral (PID) controller to adapt to changing conditions, such as different payloads in unmanned aerial vehicles \cite{vitale2021event, glatz2019adaptive, stagsted2020event, stagsted2020towards}, or to prevent drift in non-neutral buoyancy blimps~\cite{burgers2023evolving}. Much of this work successfully deployed SNNs as PID controllers in real-world systems on neuromorphic hardware, highlighting the potential for low-power autonomous control. Moving from the sky to the ground, SNN-based PID controllers have been used for lane-keeping tasks in simulated environments with reference trajectories provided by the lane~\cite{bing2018end, halaly2023autonomous, kaiser2016towards}, as well as with LiDAR for collision avoidance in simulated environments~\cite{shalumov2021lidar}.
These tasks all show successful use of SNNs in adaptive control, though the objective of a PID controller is to maintain a desired setpoint which is often a well-defined and simpler goal than end-to-end autonomous driving in the face of complex and noisy environments. We push the frontier of what SNNs are capable of in this paper.


\textbf{End-to-end Autonomous Driving} directly maps sensory inputs to vehicle control outputs using a single, fully differentiable model. Existing approaches can be broadly classified into two categories: imitation learning and reinforcement learning paradigms~\cite{chen2023end}. Imitation learning methods, such as behavior cloning~\cite{chen2020learning,bojarski2016end,codevilla2018end,hawke2020urban,codevilla2019exploring} and inverse optimal control~\cite{zeng2019end,sadat2020perceive,wang2021end,hu2021safe,khurana2022differentiable}, learn a driving policy by mimicking expert demonstrations. On the other hand, reinforcement learning techniques~\cite{kendall2019learning,liang2018cirl,toromanoff2020end,chekroun2021gri} enable the driving agent to learn through interaction with the environment by optimizing a reward function. Recent advancements, such as multi-modal sensor fusion~\cite{prakash2021multi,chitta2022transfuser,shao2022safety,jia2023think,jaeger2023hidden}, attention mechanisms~\cite{prakash2021multi,chitta2021neat,chitta2022transfuser}, and policy distillation~\cite{chen2020learning,chen2021learning,zhang2021end,wu2022trajectory,zhang2023coaching} have significantly improved the performance of end-to-end driving systems.

\begin{figure}
    \centering
    \includegraphics[width=\textwidth]{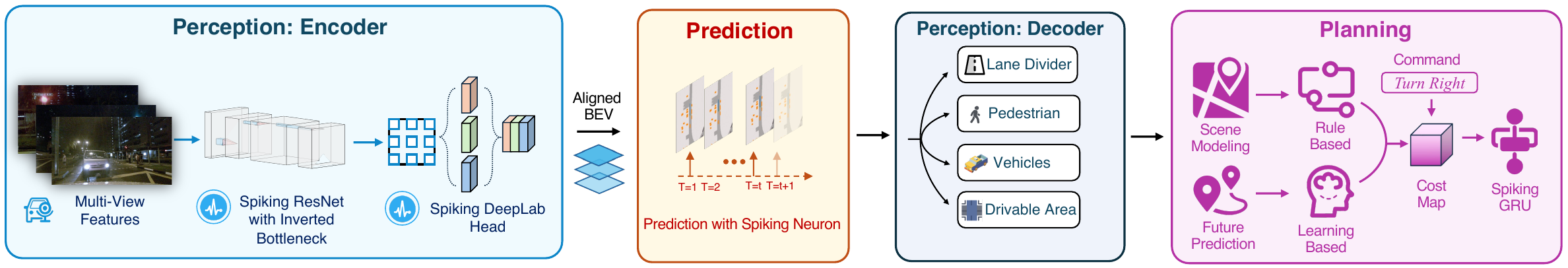}
    \caption{Overview of \name{}. The multi-view features from the perception encoder, including a spiking ResNet with inverted bottleneck and spiking DeepLab head, are fed into a prediction module using spiking neurons. The perception decoder then generates lane divider, pedestrian, vehicle and drivable area predictions. Finally, the planning module models the scene and generates future predictions to inform rule-based command decisions for turning, stopping, and goal-directed navigation.}
    \label{fig:overview}
\end{figure}

\section{Method}

This section presents the Spiking Autonomous Driving (\name{}) method, an end-to-end framework that integrates perception, prediction, and planning using SNNs (Fig.~\ref{fig:overview}). \name{}'s biologically-inspired architecture enables efficient spatiotemporal processing for autonomous driving, with the spiking neuron layer at its core. This layer incorporates spatiotemporal information and enables spike-driven computing, making it well-suited for the dynamic nature of autonomous driving tasks.

The perception module is the first stage of the \name{} framework. It constructs a bird's eye view (BEV) representation from multi-view camera inputs, providing a human-interpretable understanding of the environment. This representation serves as the foundation for the subsequent prediction and planning modules.
The prediction module uses the BEV to forecast future states using a `dual pathway',
which allows data to flow through two separate paths, providing a pair of alternative data embeddings. One pathway focuses on encoding information from the past, while the other pathway specializes in predicting future information. Subsequently, the embeddings from these two pathways are fused together, integrating the past and future information to facilitate temporal mixing. 
This enables the anticipation of dynamic changes in the environment, which is crucial for safe and efficient autonomous driving.
Leveraging the perception and prediction outcomes, the planning module generates safe trajectories by considering predicted occupancy of space around the vehicle, traffic rules, and ride comfort. To optimize the entire pipeline, \name{} is trained end-to-end using a composite loss that combines objectives from perception, prediction, and planning. The following subsections describe each module in detail.

\subsection{Spiking Neuron Layer}\label{Sec_spike_layer}
All modules consist of spiking neurons rather than artificial neurons, and so a formal definition of spiking neurons is provided below. Spiking neuron layers integrate spatio-temporal information into the hidden state of each neuron (membrane potential) which are converted into binary spikes emitted to the next layer. Spiking neurons can be represented as recurrent neurons with binarized activations and a diagonal recurrent weight matrix such that the hidden state of a neuron is isolated from all other neurons (see \cite{eshraghian2023training} for a derivation). We adopt the standard Leaky Integrate-and-Fire (LIF) \cite{maass1997networks} model, whose dynamics are described by the following equations:
\begin{align}
&U[t]=H[t-1]+X[t], \\ \label{eq_sn_layer}
&S[t]=\Theta\left(U[t]-u_{th}\right), \\
&H[t]=U_{\rm reset}S[t] + \left(\beta U[t]\right)\left(1-S[t]\right),
\end{align}
where $X[t]$ is the input to the neuron at time-step $t$, and is typically generated by convolutional or dense operators. $U[t]$ denotes the membrane potential of the neuron, and integrates $X[t]$ with the temporal input component $H[t-1]$. $\Theta(\cdot)$ is the Heaviside step function, which is 1 for $x\geq0$ and 0 otherwise. If $U[t]$ exceeds the firing threshold $u_{th}$, the spiking neuron emits a spike $S[t]=1$ as its activation, and the temporal output $H[t]$ is reset to $V_{\rm reset}$. Otherwise, no spike is emitted ($S[t]=0$) and $U[t]$ decays to $H[t]$ with a decay factor $\beta < 1$. 
For brevity, we refer to Eq.~\ref{eq_sn_layer} as ${\mathcal{SN}}(\cdot)$, where the input $U$ is a tensor of membrane potential values fed into multiple spiking neurons, and the output $S$ is an identically-shaped tensor of spikes.

\subsection{Perception: Distinct Temporal Strategies for Encoder and Decoder
} \label{sec:perception}
\begin{figure}
    \centering
    \includegraphics[width=\linewidth]{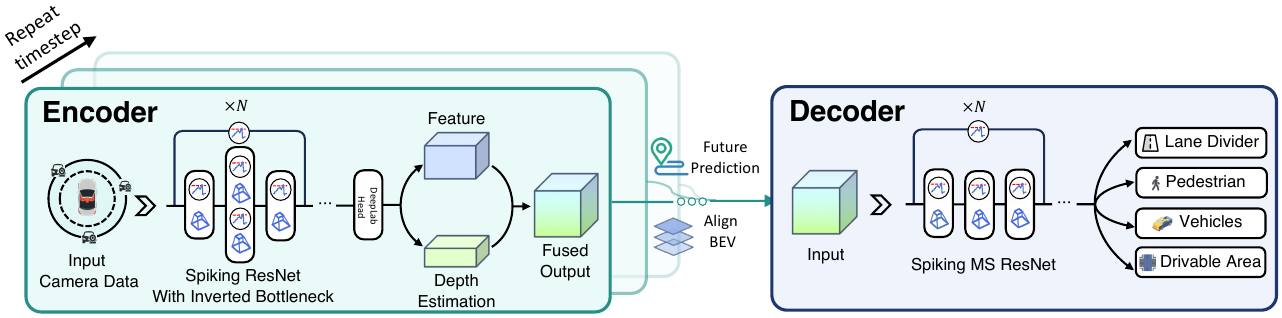}
    \caption{The perception module. The encoder takes multi-camera input data, passes it through a spiking ResNet with inverted bottleneck to generate feature representations, each of which has its own depth estimation. These are fused and passed to the decoder, which generates predictions for lane dividers, pedestrians, vehicles and drivable areas.}
    \label{fig:perception}
\end{figure}

Fig.~\ref{fig:perception} illustrates the overall architecture of the perception module. The perception stage constructs a spatiotemporal BEV representation from multi-view camera inputs over $t$ time-steps through spatial and temporal fusion of features extracted from the cameras. It consists of an encoder, which processes each camera input to generate features and depth estimations, and a decoder, which generates BEV segmentation and instructs the planning module. A future prediction module is depicted between the encoder and decoder in Fig.~\ref{fig:perception}. It is not used in the first stage where the perception module is trained alone, but it is included in the second stage once the prediction module is included.


The temporal dimension processing in the Encoder/Decoder architecture is a crucial design consideration, as both SNNs and autonomous driving data inherently possess a temporal structure. There are two approaches to handle this:

\begin{itemize}
    \item \textbf{Sequential Alignment (SA):} sequential input data is passed to the SNN step-by-step by aligning the time-varying dimension of the input data with the model
    \item \textbf{Sequence Repetition (SR):} sequential input data is aligned with the batch dimension for better parallelism during training, and individual frames are repeated $T$ times over the model sequence, so as to create virtual timesteps. SR is commonly used for pre-training sequence-based models on static image datasets. 
\end{itemize}



Given these two encoding options, we test all four combinations of these options applied to both the encoder and decoder of the perception block. Based on our experiments (detailed in Sec.~\ref{sec:abl-timestep}), the best performing approach is using SR for the encoder and SA for the decoder. The encoder is also pre-trained on ImageNet-1K which requires the use of repeated images to create virtual timesteps. Further details regarding the pre-training of the encoder can be found in Appendix~\ref{appendix:pretrain}. Conversely, the decoder is trained from scratch, which naturally assumes a temporal-mixing role, making the alignment of sequential data with the model sequence a more effective approach.


The training process involves first training the encoder-decoder, followed by the prediction module. This approach integrates spatial and temporal information for comprehensive BEV representation in autonomous vehicle perception and planning.

\paragraph{Encoder: Spiking Token Mixer with Sequence Repetition
}
The encoder module can be thought of as a spiking token mixer (STM). The STM consists of 12-layers of spiking CNN pre-trained on ImageNet-1K~\cite{deng2009imagenet} to generate vision patch embeddings, which is effectively a deeper version of the `spiking patch embedding' from Ref.~\cite{yao2024spike,zhou2022spikformer}.
Across these 12 layers, the number of channels in each layer is designed to first increase and then decrease so as to act as an inverted bottleneck. While SPS layers are usually terminated by self-attention, we replaced this with dense layers instead, which both reduces computational resources and leads to improved performance.
In doing so, we achieved a 72.1\% ImageNet top-1 classification accuracy with only 12M parameters. In contrast, the previous spiking vision transformers that employs self-attention reached 70.2\% with the same number of parameters~\cite{yao2024spike}.

The encoder extracts feature embeddings and depth estimations while squeezing the image into a smaller latent space. The overall workflow of the encoder can be summarized as follows:
\begin{equation*}
\begin{aligned}
&X = \text{STM}(I), && I \in \mathbb{R}^{N \times C_{in} \times T \times L \times H \times W}, \quad X \in \mathbb{R}^{C \times T \times L \times H \times W} \\
&\mathcal{F} = \text{Head}_{\rm feature}(X), && X \in \mathbb{R}^{C \times T \times L \times H \times W}, \quad \mathcal{F} \in \mathbb{R}^{C_f \times L \times H \times W}, \\
&\mathcal{D} = \text{Head}_{\rm depth}(X), && X \in \mathbb{R}^{C \times T \times L \times H \times W}, \quad \mathcal{D} \in \mathbb{R}^{C_d \times L \times H \times W} \\
&Y = \mathcal{F} \otimes \mathcal{D}, &&Y \in \mathbb{R}^{C_f \times C_d \times T \times L \times H \times W}\\
\end{aligned}
\end{equation*}

The STM encoder described above is used to extract feature embeddings and depth estimates from each camera frame $I_t \in \mathbb{R}^{N \times C_{in} \times H \times W}$, where $N=6$ is the number of cameras, $C_{in}=3$ refers to the number of input channels (RGB), and $H \times W$ refers to the video resolution. Note that the use of sequence repetition means that $T$ is the number of times the same frame is repeated over the sequence, while $L$ is the number of frames in a continuous camera recording. As such, the dimensions $N \times L$ are stacked so as to speed up processing.


The encoder consists of 12 layers, each containing a 2D convolution layer, batch normalization, and spiking neuron. The output of the encoder is a feature map $\mathcal{F}\in \mathbb{R}^{C_f \times L \times H \times W} $ and a depth estimation $\mathcal{D}\in \mathbb{R}^{C_d \times L \times H \times W}$, where $C_f$ is the number of feature channels, $C_d$ is the number of channels, each of which has a depth associated with it, and $(H, W)$ is the spatial size.
Formally, given an image sequence ${{I}}$:
\begin{align}
X={\mathcal{SN}}({\rm{BN}}({\rm{Conv2d}}(I))),
\end{align}
where Conv2d represents a 2D convolutional layer (stride: 1, $3 \times 3$ kernel size),  $BN$ is batch normalization, and $\mathcal{MP}$ is a max-pooling operator.
The feature map  $\mathcal{F}$ and depth estimation $\mathcal{D}$ are then averaged over the sequence $T$ and combined using an outer product to obtain a camera feature frustum:
\begin{equation}
Y = \mathcal{F} \otimes \mathcal{D}, Y \in \mathbb{R}^{C_f \times C_d \times T \times L \times H \times W}
\end{equation}
The frustums from all cameras are transformed into a global 3D coordinate system centered at the ego-vehicle's inertial center at time $i$. Previous BEV feature maps are merged with the current BEV by applying a discount factor $\alpha$ to integrate these layers efficiently.
\begin{equation}
\tilde{x}_t = b_t + \sum_{i=1}^{t-1} \alpha^i \times \tilde{x}_{t-i}
\end{equation}
where $\tilde{x}_t$ is the BEV at time $t$ which has an initial condition of $\tilde{x}_1 = b_1$, and the discount factor is $\alpha = 0.5$.
We then average across the $T$ dimension to eliminate the repeated temporal dimension and obtain the average firing rate. The resulting feature map is then be passed to the decoder.




\paragraph{Decoder: Sequential Alignment with Streaming Feature Maps}

The recurrent decoder aligns feature maps sequentially, introducing a new instance of data at each time-step, contrasting with the encoder's repeated inputs. Using SA rather than SR for the decoder improves performance for two reason: 1) the decoder does not need to be pre-trained on static, repeated data, and 2) the decoder acts as a temporal mixer. In this architecture, time-mixing is achieved using LIF neurons and allows them to take on the role of self-attention without the same computational burden. The LIF neurons are composed as a shared backbone as a set of layers used to extract features from data before being passed to various specialized heads, each dedicated to a specific task. The high-level dataflow summarized as follows:

\begin{equation*}
\begin{aligned}
&X = \text{SharedBackbone}(I_D), && I \in \mathbb{R}^{C_{in} \times L \times H \times W}, \quad X \in \mathbb{R}^{C_{med} \times T \times L \times H \times W} \\
&Y_k = \text{Head}_k(X), && Y_k \in \mathbb{R}^{C_{out} \times L \times H \times W}, \quad k \in \{\text{seg, ped, map, inst}\} \\
\end{aligned}
\end{equation*}

where $I_D$ is the input tensor to the decoder with dimensions $(C_{in}, L, H, W)$, $X$ is the output of the shared backbone with dimensions $(C_{med}, T, L, H, W)$,$Y_k$ is the output of the $k$-th head with dimensions $(C_{out}, L, H, W)$, and $k$ indexes into the heads of the different tasks: vehicle segmentation (seg), pedestrian (ped), HD map (map), and future instance (inst). The shared backbone is implemented using the first three layers of MS-ResNet18~\cite{hu2024advancing} followed by three upsampling layers with a factor of 2 and skip connections. More details about MS-ResNet can be found in Appendix~\ref{appendix:ResNet}. The resulting features have 64 channels and are then passed to different heads according to the task requirements. Each head consists of a spiking convolutional layer.




\subsection{Prediction: Fusing Parallel Spike Streams
}\label{sec: method-prediction}

Predicting future agent behavior is crucial for an autonomous vehicle to be reactive and make informed decisions in real-time. In our approach, we accumulate historical BEV features and predict the next few timesteps using purely LIF neurons. However, the stochastic nature of interactions among agents, traffic elements and road conditions, makes it challenging to accurately predict future trajectories. To address this, we model future uncertainty with a conditional Gaussian distribution.

To accomplish this, two layers of LIF neurons are used. 
The first parallel layer takes the present and prior output BEV feature maps from the encoder of the perception model  $(x_1, \dots, x_t)$ as inputs. The first BEV $x_1$ is also used as the initial membrane potential for this LIF layer.
The second parallel layer accounts for the uncertainty distribution of future BEV predictions. The uncertainty distribution is generated by passing the present feature $x_t$ through 4 spiking MS-ResNet blocks, average pooling, and another 2D spiking convolution with a kernel size of $(1,1)$ to transform the channel depth to double the output of the first parallel layer. Another averaging pooling operator compresses the feature map down to a vector. The vector is split in two sub-vectors representing mean $\mu \in \mathbb{R}^L$ and variance $\sigma \in \mathbb{R}^L$, and these values populate a diagonal Gaussian distribution of the latent feature map. Using the $\mu$ and standard deviation $\sigma$, we can construct the Gaussian distribution at timestep $t$, denoted as $\eta_t$. This distribution, represented by the parameters $\mu$ and $\sigma$, can then be concatenated with the input at the current timestep $x_t$.


Simultaneously, all prior spiking outputs are concatenated to the present input $x_t$, denoted below $x_{0:t}$. The predicted BEV feature for the next timestep is calculated below:

\begin{equation}
\hat{x}_{t+1} = \mathrm{LIF}({\rm{BN}}({\rm{Conv2d}}(\text{concatenate}(x_t, \eta_t))) \oplus \mathrm{LIF}({\rm{BN}}({\rm{Conv2d}}((x_{0:t}))),
\end{equation}
where $\hat{x}_{t+1}$ represents the predicted BEV features for the next timestep, $\mathrm{LIF}(\cdot)$ denotes the LIF neuron layer, $\text{concatenate}(\cdot)$ represents the concatenation operation, and $\oplus$ denotes element-wise addition of the outputs from the two LIF layers, and the inner $\mathrm{Conv2d}(\cdot)$ and $\mathrm{BN}(\cdot)$ layers are used to ensure consistency in the output dimensions of the first and second LIF layers.
\begin{wrapfigure}{r}{0.5\textwidth}

  \begin{center}
    \includegraphics[width=0.48\textwidth]{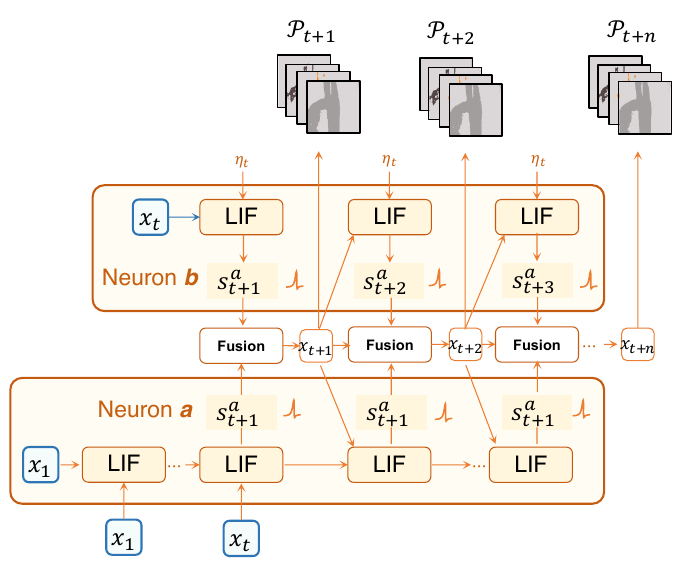}
    \
  \end{center}
  
  \caption{Dual pathway modeling for prediction. Neuron $a$ captures future multi-modality by incorporating uncertainty distribution. Neuron $b$ compensates for information gaps using past variations. Inputs $x_1$ and $x_t$ from both pathways are used for the next prediction step.}
  \label{fig:prediction}
\end{wrapfigure}
The mixture prediction serves as the basis for the subsequent prediction steps. By recursively applying this dual-pathway prediction method, we obtain the predicted future states $(\hat{x}_{t+1},\dots, \hat{x}_{t+n})$. The overall datapath is illustrated in Fig.~\ref{fig:prediction}.
Following the dual-pathway prediction, all the features are fed into a Spiking ResNet using SA for additional temporal mixing. The historical features $(x_1, \dots, x_t)$ and the predicted future features $(\hat{x}_{t+1},\dots, \hat{x}_{t+n})$ are then passed to the perception decoder, which consists of multiple output heads to generate various interpretable intermediate representations.

\subsection{Planning: Temporal Spiking Temporal for Trajectory Refinement} \label{sec:planning}

The primary objective of the \name{} system is to plan a trajectory that is both safe and comfortable, aiming towards a designated target. This is accomplished through a planning pipeline that integrates predicted occupancy grids $o$, map representations $m$, and temporal dynamics of environmental elements like pedestrians.

\textbf{Motion Planning.} Initially, our motion planner generates a diverse set of potential trajectories using the bicycle model~\cite{polack2017kinematic}. Among these, the trajectory that minimizes a predefined cost function is selected. This function integrates various factors including learned occupancy probabilities (from segmentation maps generated in the Prediction section) and compliance with traffic regulations to ensure the selected trajectory optimizes for safety and smoothness.

\textbf{Cost Function.} The cost function $f$ employed is a multi-component function, where:
\begin{equation}
    f(\tau, o, m; w) = f_o(\tau, o, m; w_o) + f_v(\tau; w_v) + f_r(\tau; w_r)
    \label{eq:obj}
\end{equation}
where $w = (w_o, w_v, w_r)$ represents the learnable parameters associated with each cost component, and $\tau$ denotes a set of trajectory candidates. Specifically, $f_o$ evaluates trajectory compliance with static and dynamic obstacles, $f_v$ is derived from the prediction decoder assessing future states, and $f_r$ addresses metrics of ride comfort and progress towards the goal. The aim is to select the optimal set of trajectories $\tau^*$ that minimize this cost function. This cost function is adapted from Hu et al.~\cite{hu2022st} and is detailed in Appendix~\ref{appendix:cost}.

Additionally, trajectories are filtered based on high-level commands (e.g., go forward, turn left, turn right) which tailor the trajectory selection to the immediate navigational intent.

\textbf{Optimization with Spiking Gated Recurrent Unit.} Following the initial selection, the "best" trajectory $\tau^*$  undergoes further refinement using a Spiking Gated Recurrent Unit (SGRU), as inspired by~\cite{lotfi2020long}. In this optimization phase, the hidden state $h_t$ of the SGRU incorporates features derived from the front camera's encoder. The input state $x_t$ is formulated by concatenating the vehicle's current position, the corresponding position from the selected trajectory $\tau^*$, and the designated target point.

The SGRU model processes inputs as follows:
\begin{align}
    r_t &= \Theta(W_{ir} x_t + b_{ir} + W_{hr} h_{t-1} + b_{hr}) \\
    z_t &= \Theta(W_{iz} x_t + b_{iz} + W_{hz} h_{t-1} + b_{hz}) \\
    n_t &= \Theta(W_{in} x_t + b_{in} + r_t \odot (W_{hn} h_{t-1} + b_{hn})) \\
    h_t &= (1 - z_t) \odot n_t + z_t \odot h_{t-1}
\end{align}
where $h_t$ denotes the hidden state at time $t$, $x_t$ represents the input, and $r_t$, $z_t$, $n_t$ are the reset, update, and new candidate state gates respectively. The Heaviside function $\Theta$ ensures sparse and binarized operations in the state of the SGRU, thus preserving the advantages of SNNs.

This optimization step enhances trajectory reliability by mitigating uncertainties inherent in perceptual and predictive analyses, and by integrating dynamic traffic light information directly into the trajectory planning process.

\subsection{Overall Loss for End-to-End Learning}\label{sec:method-e2e_learning}

Our model integrates perception, prediction, and planning into a unified framework optimized end-to-end using a composite loss function:
\begin{equation}
    \mathcal{L} =  \mathcal{L}_{per} + \alpha \mathcal{L}_{pre} + \beta \mathcal{L}_{pla}
\end{equation}
where $\alpha$ and $\beta$ are learnable weights. These parameters dynamically adjust to scale the contributions of each task based on the gradients from the respective task losses.

\noindent\textbf{Perception Loss.}
This component is multi-faceted, covering segmentation for current and previous frames, mapping, and auxiliary depth estimation. For semantic segmentation, a top-k cross-entropy loss is employed to effectively handle the large amount of background, non-salient content in the BEV images, following the approach in~\cite{hu2021fiery}. Instance segmentation utilizes an $l_2$ loss for centerness supervision and an $l_1$ loss for offsets and flows. Lane and drivable area predictions are evaluated using a cross-entropy loss. 

\noindent\textbf{Prediction Loss.}
Our prediction module extends the perception task by forecasting future semantic and instance segmentation. It adopts the same top-k cross-entropy loss used in perception but applies an exponential discount to future timestamps to account for increasing uncertainty in predictions.

\noindent\textbf{Planning Loss.}
The planning module begins by selecting the initial best trajectory $\tau^*$ from a set of sampled trajectories as defined in Eq.~\ref{eq:obj}. This trajectory is then refined using an SGRU-based network to produce the final trajectory output $\tau_o^*$. The planning loss comprises two parts: a max-margin loss that differentiates between the expert behavior $\tau_h$ (treated as a positive example) and other sampled trajectories (treated as negatives), and an $l_1$ distance loss that minimizes the deviation between the refined trajectory and the expert trajectory.

Further details on these loss components and the training are provided in the Appendix~\ref{appendix:stage_training}.

\newcommand{\cmark}{\raisebox{0pt}{\color{ForestGreen}\ding{51}}}%
\newcommand{\xmark}{\raisebox{0pt}{\color{Red}\ding{55}}}

\begin{table}[t!]
\caption{\textbf{Perception results.} We report the BEV segmentation IoU (\%) of intermediate representations and their mean value.}
\label{tab:perception}
\centering
\begin{tabular}{p{0.22\textwidth}|>{\centering}p{0.05\textwidth}|>{\centering}p{0.1\textwidth}>{\centering}p{0.1\textwidth}>{\centering}p{0.1\textwidth}>{\centering}p{0.1\textwidth}>{\centering\arraybackslash}p{0.1\textwidth}}
\toprule
\multirow{2}{*}{Method}   & \multirow{2}{*}{Spike}  &Drivable Area     & \multirow{2}{*}{Lane}     & \multirow{2}{*}{Vehicle}     & \multirow{2}{*}{Pedestrian}     & \multirow{2}{*}{Avg.}     \\ \midrule
VED~\cite{lu2019monocular}\textsuperscript{RAL}  &  \xmark    & 60.82 & 16.74 & 23.28 & 11.93 & 28.19 \\
VPN~\cite{pan2020cross}\textsuperscript{RAL}   & \xmark                     & 65.97 & 17.05 & 28.17 & 10.26 & 30.36 \\
PON~\cite{roddick2020predicting}\textsuperscript{CVPR}  & \xmark     & 63.05 & 17.19 & 27.91 & 13.93 & 30.52 \\
Lift-Splat~\cite{philion2020lift}\textsuperscript{ECCV} &\xmark & 72.23 & 19.98 & 31.22 & 15.02 & 34.61 \\
IVMP~\cite{wang2021learning}\textsuperscript{ICRA}  &  \xmark   & 74.70  & 20.94 & 34.03 & 17.38 & 36.76 \\
FIERY~\cite{hu2021fiery}\textsuperscript{ICCV}  &  \xmark  & 71.97 & 33.58 & 38.00 & 17.15 & 40.18 \\ 
ST-P3~\cite{hu2022st}\textsuperscript{ECCV}    & \xmark  &75.97       &33.85       & 38.00       &17.15       &42.69       \\ 
\midrule
\rowcolor{mygray}\textbf{\name{ (Ours)}} & \cmark& 64.74 & 27.78 & 34.82 & 15.12  & 35.62\\

\bottomrule
\end{tabular}
\end{table}

\section{Experiments}

We evaluate the proposed model using the nuScenes dataset~\cite{caesar2020nuscenes} with 20 epochs with SpikingJelly~\cite{fang2023spikingjelly} framework. For our experiments, we consider $1.0s$ of historical context and predict $2.0s$ into the future, which corresponds to processing 3 past frames and predicting 4 future frames. Additional experimental details are elaborated in the Appendix.

\subsection{Experimental Results on nuScenes}\label{sec:res-nuscenes}

\textbf{Perception.} Our evaluation focuses on the model's ability to interpret map representations and perform semantic segmentation. For map representation, we specifically assess the identification of drivable areas and lanes, which are critical for safe navigation as they dictate where the Self-Driving Vehicle (SDV) can travel and help maintain the vehicle's position within the lanes. Semantic segmentation tests the model's ability to recognize dynamic objects, such as vehicles and pedestrians, which are pivotal in urban driving scenarios.

We employ the Intersection-over-Union (IoU) metric to quantify the accuracy of our BEV segmentation tasks. The results, as summarized in Tab.~\ref{tab:perception}, show that our SAD method, which is fully implemented with spiking neural networks (SNNs), competes favorably against state-of-the-art, non-spiking artificial neural networks (ANNs). Notably, our model achieves a superior mean IoU on the nuScenes dataset compared to existing leading methods such as VED~\cite{lu2019monocular}, VPN~\cite{pan2020cross}, PON~\cite{roddick2020predicting}, and Lift-Splat~\cite{philion2020lift}. Specifically, our SAD method outperforms the VED~\cite{lu2019monocular} model by \textbf{7.43\%} in mean IoU. This enhancement is significant, considering that our network utilizes spiking neurons across all layers, which contributes to greater computational efficiency. Despite the inherent challenges of using SNNs, such as the binary nature of spikes and potential information loss compared to traditional ANNs, our results demonstrate that SAD is capable of delivering competitive perception accuracy in autonomous driving scenarios.

\begin{table}[t!]
\caption{\textbf{Prediction results.} We report semantic segmentation IoU (\%) and instance segmentation metrics from the video prediction area. The \textit{static} method assumes all obstacles static in the prediction horizon.}
\label{tab:prediction}
\centering
\begin{tabular}{>{\raggedright}p{0.22\textwidth}|>{\centering}p{0.05\textwidth}|>{\centering}p{0.22\textwidth}|>{\centering}p{0.1\textwidth}>{\centering}p{0.1\textwidth}>{\centering\arraybackslash}p{0.1\textwidth}}
\toprule
\multicolumn{1}{l|}{\multirow{2}{*}{Method}} & \multirow{2}{*}{Spike} & Future Semantic Seg. & \multicolumn{3}{c}{Future Instance Seg.} \\
\multicolumn{1}{c|}{}                        &                        &                  IoU $\uparrow$                                    & PQ $\uparrow$ & SQ $\uparrow$ & RQ $\uparrow$ \\ \midrule
Static                                       &  \xmark                & 32.20                                                & 27.64         & 70.05         & 39.08         \\
FIERY~\cite{hu2021fiery}\textsuperscript{ICCV} & \xmark               & 37.00                                                & 30.20         & 70.20         & 42.90         \\ 
ST-P3~\cite{hu2022st}\textsuperscript{ECCV}   & \xmark               & 38.63                                                & 31.72         & 70.15         & 45.22         \\ \midrule
\rowcolor{mygray}
\textbf{\name{ (Ours)}}                        &  \cmark              & 32.74                                                & 20.00         & 68.74         & 29.39         \\
\bottomrule
\end{tabular}
\end{table}

\begin{table}[tb!]
\centering
\caption{\textbf{Planning results.}}
\label{tab:planning}
\begin{tabular}{p{0.22\textwidth}|>{\centering}p{0.05\textwidth}|>{\centering}p{0.05\textwidth}>{\centering}p{0.05\textwidth}>{\centering}p{0.05\textwidth}|>{\centering}p{0.05\textwidth}>{\centering}p{0.05\textwidth}>{\centering\arraybackslash}p{0.05\textwidth}|>{\centering\arraybackslash}p{0.13\textwidth}}
\toprule
\multirow{2}{*}{Method} & \multirow{2}{*}{Spike} & \multicolumn{3}{c|}{L2 ($m$) $\downarrow$} & \multicolumn{3}{c|}{Collision (\%) $\downarrow$} & \multirow{2}{*}{Energy (mJ)} \\ \cline{3-8} 
                        &                        & 1s     & 2s    & 3s    & 1s        & 2s        & 3s   &     \\ \midrule

NMP~\cite{zeng2019end}\textsuperscript{CVPR}       & \xmark              & 0.61   & 1.44  & 3.18  & 0.66      & 0.90      & 2.34  &  -   \\
Freespace~\cite{hu2021safe}\textsuperscript{CVPR}  & \xmark              & 0.56   & 1.27  & 3.08  & 0.65      & 0.86      & 1.64   &  344.11 \\ 
ST-P3~\cite{hu2022st}\textsuperscript{ECCV}        & \xmark              & 1.33   & 2.11  & 2.90  & 0.23      & 0.62      & 1.27  & 3520.40\\\midrule
\rowcolor{mygray}\textbf{\name{ (Ours)}}      & \cmark              & 1.53   & 2.35  & 3.21  & 0.62      & 1.26      & 2.38 & 46.92    \\ 
\bottomrule
\end{tabular}
\end{table}

\noindent\textbf{Prediction.}
We assess the predictive capabilities of our model using multiple metrics tailored for video prediction, specifically IoU, existing Panoptic Quality (PQ), Recognition Quality (RQ), and Segmentation Quality (SQ) for evaluating our prediction quality. The definition of these metrics can be found in Appendix.~\ref{appendix:metrics}. The results, presented in Tab.~\ref{tab:prediction}, demonstrate that while our model does not employ an additional temporal module, the inherent temporal dynamics of spiking neurons facilitate effective information processing. However, our SAD model still shows a gap in performance when compared with state-of-the-art ANN methods.

\noindent\textbf{Planning.}
In the planning domain, we evaluate our model using two primary metrics: L2 error and collision rate. To ensure fairness, the planning horizon is adjusted to $3.0s$. The L2 error measures the deviation between the SDV’s planned trajectory and the human driver’s actual trajectory, providing a quantitative measure of planning accuracy. The collision rate assesses the model’s ability to safely navigate the driving environment without incidents. Results, detailed in Tab.~\ref{tab:planning}, reveal that our SAD method achieves an L2 error and collision rate comparable to those of the state-of-the-art ANN-based methods, underscoring the safety and reliability of our planning approach.

\noindent\textbf{Energy Efficiency.} Neuromorphic hardware is able to take advantage of small activation bit-widths and dynamical sparsity~\cite{davies2018loihi}, and as such, SNNs are able to significantly reduce energy consumption during inference $-$ provided there are sufficiently sparse activation patterns amongst spiking neurons.
As detailed in Table~\ref{tab:planning}, we present an estimation of the energy usage of each SOTA model based on dynamical sparsity (detailed calculation methods described in Appendix~\ref{appendix:energy}). Owing to the utilization of spiking neurons, our model achieves substantial energy reductions: \textbf{7.33 $\times$} less than the Freespace model~\cite{hu2021safe} and \textbf{75.03 $\times$} lower compared to the ST-P3 model~\cite{hu2022st}. This exceptional energy efficiency makes our model highly suitable for real-world applications.

\begin{table}[t!]
\centering
\caption{Ablation study on different timestep alignment strategies for the encoder and decoder on perception tasks. `SR' denotes repeating the timestep input, `SA' indicates aligning the timestep with the model's inherent temporal dimension, and 'w/o T' means eliminating the decoder's inherent timestep association, resulting in no hidden state connections between timesteps. Our final model adopts the configuration of repeating the timestep in the encoder and aligning the timestep in the decoder.}
\label{tab:abl_ts}
\setlength\tabcolsep{8pt} 
\renewcommand{\arraystretch}{1}
\begin{tabular}{@{}cc|ccc|ccccc@{}}
\toprule
\multicolumn{2}{c|}{\textbf{Encoder}} & \multicolumn{3}{c|}{\textbf{Decoder}} & \multicolumn{5}{c}{\textbf{Results}} \\ \midrule
\textbf{SR} & \textbf{SA} & \textbf{SR} & \textbf{SA} & \textbf{w/o T} & \textbf{Drivable} & \textbf{Lane} & \textbf{Vehicle}& \textbf{Pedestrian} & \textbf{Avg.} \\
\midrule
 &\cmark  &  &\cmark  &  &43.84 &15.25 & 4.41 & 1.62& 16.28 \\
\cmark &  &\cmark  &  &  &0.00  &0.00 & 0.00&0.42 & 0.11\\
\cmark &  &  &  & \cmark &61.80 &20.92 &31.78 & 13.46& 31.99 \\ 
\cmark & &  &\cmark  &  &\textbf{61.81} &\textbf{25.31} & \textbf{33.89} & \textbf{15.24} & \textbf{34.06} \\ \bottomrule
\end{tabular}
\end{table}

\subsection{Visualization}
\begin{figure}
    \centering
    \includegraphics[width=\linewidth]{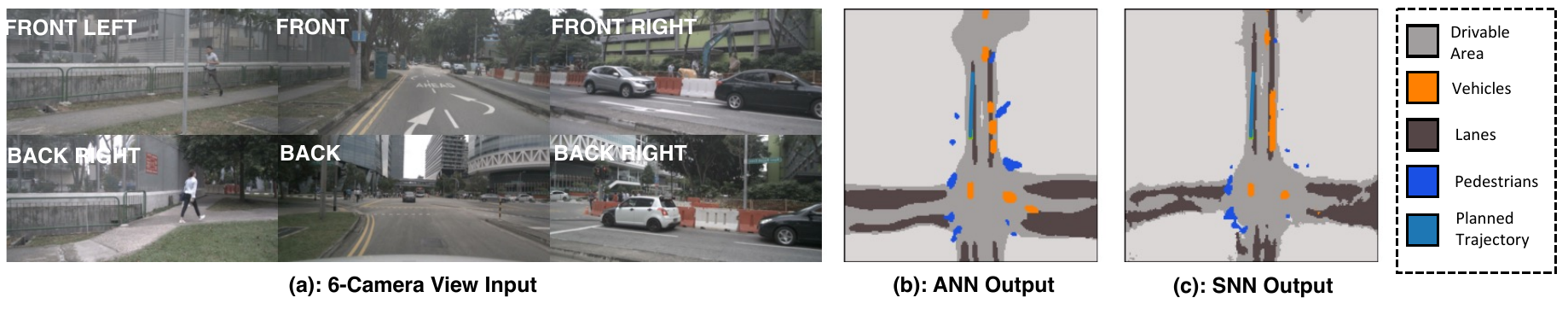}
    \caption{\textbf{Qualitative Results of the SAD Model on the nuScenes Dataset.} (a) displays six camera view inputs utilized by the model. (b) illustrates the planning result of the ANN model, and (c) presents the planning results of our SAD model. The comparison shows that our SAD model can achieve performance comparable to that of the ANN model and successfully generate a safe trajectory.}
    \label{fig:visual1}
\end{figure}

To evaluate the effectiveness of our SAD model in a more interpretable manner, we provide qualitative results using the nuScenes dataset. Fig.~\ref{fig:visual1} illustrates the outputs of our model. The SAD model effectively generates a safe trajectory that enables straight-ahead motion while avoiding collisions with curbsides and the vehicle ahead. Additionally, we conducted a comparative analysis with the ANN model, which demonstrated that our SAD model can achieve comparable performance, ensuring accurate and reliable planning outcomes.More visual results can be found in the Appendix~\ref{appendix:visual}.

\subsection{Ablation Study}

\subsubsection{Timestep Strategy} 
\label{sec:abl-timestep}
The large space of architecture design decisions came with a large number of ablation studies before determining the best performing model.
We conducted an ablation study to examine the effects of different timestep alignment strategies on the performance of the encoder and decoder in perception tasks. The strategies include `SR' for the repetition of vision inputs over timesteps, `SA' where sequential inputs are aligned with the model's inherent temporal dimension, and 'w/o T' where recurrent dynamics in the decoder are removed, thus disconnecting the hidden states between timesteps.
The results are summarized in Tab.~\ref{tab:abl_ts}. The last row is the baseline configuration of our model that serves as a reference point for these experiments. All ablation experiments are run for 5 epochs.

\textbf{Impact of Timestep Repetition in Encoder.} (Row 1) 
When repeating timesteps in the encoder which was pre-trained on the ImageNet-1K dataset for classification tasks, we notice a substantial benefit. 
This approach leverages the encoder's ability to capture spatial information through repeated images, a technique effective during its pre-training phase. 
Adopting this strategy in the SAD model enhances perception performance by maintaining consistency with the pre-training approach.

\textbf{Impact of Timestep Alignment in Decoder.} (Row 2) In contrast to the encoder, the decoder, which is trained from scratch, benefits from aligning timesteps with the model's inherent temporal dimension (`SA'). This strategy leverages the decoder’s capacity to mix temporal information, improving performance over the repeat strategy (`SR'), which fails to show any performance gains.

\textbf{Single Timestep Input.} (Row 3) 
The purpose of this experiment was to study how well spiking neurons perform as temporal mixers.
In this setup, all inputs are processed in one timestep without inter-frame connections, leading to a slight performance decrease compared to our baseline. Therefore, it confirms that spiking neurons inherently possess temporal processing capabilities.

\subsubsection{Effectiveness of Different Modules}

Tab.~\ref{tab:abl_md} outlines the results of an ablation study that assesses the impact of various structural modifications to the encoder and decoder on planning tasks. The study differentiates between `MS' for MS-ResNet structure~\cite{hu2024advancing}, `SEW' for SEW-ResNet structure~\cite{fang2021deep}, `SP' for single pathway model, and `DP' for dual pathway model. We analyze the planning performance of each configuration. The last row represents the configuration of our final model, which we use as the baseline for comparison.

\textbf{Decoder Structure Evaluation.} (Row 1) This part of the study compares the MS-ResNet and SEW-ResNet structures in their roles as decoders. Each structure possesses unique firing patterns, which are discussed in detail in the Appendix~\ref{appendix:ResNet}. Our results indicate that the MS-ResNet structure is more effective for planning tasks in autonomous driving, likely due to its enhanced capability to handle the spatial-temporal dynamics required in this context.

\textbf{Pathway Model Comparison.} (Row 2) Here, we explore the performance difference between single and dual pathway prediction models. The dual pathway model integrates information through two distinct processing streams. Our experimental results show that the dual pathway model significantly outperforms the single pathway model in planning tasks.

\begin{table}[h!]
\centering
\caption{Ablation Study on different modules for the encoder and the decoder on Planning tasks. `MS' denotes the  MS-ResNet structure, SEW denotes the SEW-ResNet structure, `SP' means the single pathway model, and `DP' means the dual pathway model.}
\label{tab:abl_md}
\setlength\tabcolsep{10pt} 
\renewcommand{\arraystretch}{1}
\begin{tabular}{@{}cc|cc|ccc@{}}
\toprule
\multicolumn{2}{c|}{\textbf{Decoder}} & \multicolumn{2}{c|}{\textbf{Prediction}} & \multicolumn{3}{c}{\textbf{Results}} \\ \midrule
\textbf{MS} & \textbf{SEW} &  \textbf{SP} & \textbf{DP} & \textbf{PQ} & \textbf{SQ} & \textbf{RQ} \\
\midrule
 &  \cmark &  &  \cmark&59.28  &0.75 & 0.44 \\
\cmark &  &\cmark  &   &67.55 &13.35 &19.77 \\
 \cmark&  &  &\cmark  &\textbf{68.16} &\textbf{16.57} & \textbf{24.11}  \\\bottomrule
\end{tabular}
\end{table}

\section{Conclusion}
In this work, we presented Spiking Autonomous Driving (SAD), the first end-to-end spiking neural network designed for autonomous driving. By integrating perception, prediction, and planning into a unified neuromorphic framework, SAD demonstrates competitive performance on the nuScenes dataset while exhibiting exceptional energy efficiency compared to state-of-the-art ANN-based methods. The perception module effectively constructs interpretable bird's eye view representations, the prediction module accurately forecasts future states using a novel dual-pathway architecture with spiking neurons, and the planning module generates safe and comfortable trajectories. Crucially, SAD showcases the immense potential of SNNs in complex real-world applications, marking a significant step towards realizing low-power, intelligent systems for safety-critical domains like autonomous vehicles. Moving forward, we believe this work will inspire further research into neuromorphic computing for sustainable and robust autonomous driving solutions.

\section*{Acknowledgements}
This work was supported in part by National Science Foundation (NSF) awards CNS-1730158, ACI-1540112, ACI-1541349, OAC-1826967, OAC-2112167, CNS-2100237, CNS-2120019, the University of California Office of the President, and the University of California San Diego's California Institute for Telecommunications and Information Technology/Qualcomm Institute. Thanks to CENIC for the 100Gbps networks. This project was also supported in-part by the National Science Foundation RCN-SC 2332166.
\bibliography{neurips}
\bibliographystyle{unsrt}

\newpage
\section*{\textbf{APPENDIX}}
\appendix
\section{Spiking ResNet Architecture}
\label{appendix:ResNet}
MS-ResNet~\cite{hu2024advancing} and SEW-ResNet~\cite{fang2021deep} are two directly trained residual learning methods for SNNs, primarily aiming to overcome the degradation of spiking activity in training deep SNNs. In classic CNNs, ResNet achieves "very deep" neural networks by attaching an identity mapping skip connection throughout the entire network. However, directly copying the classic residual structure to SNNs causes the training loss increases as the network deepens. The underlying reason is that CNNs generate continuous valued activations, while SNNs generate
discrete spikes.

\begin{figure}
    \centering
    \includegraphics[width=0.6\linewidth]{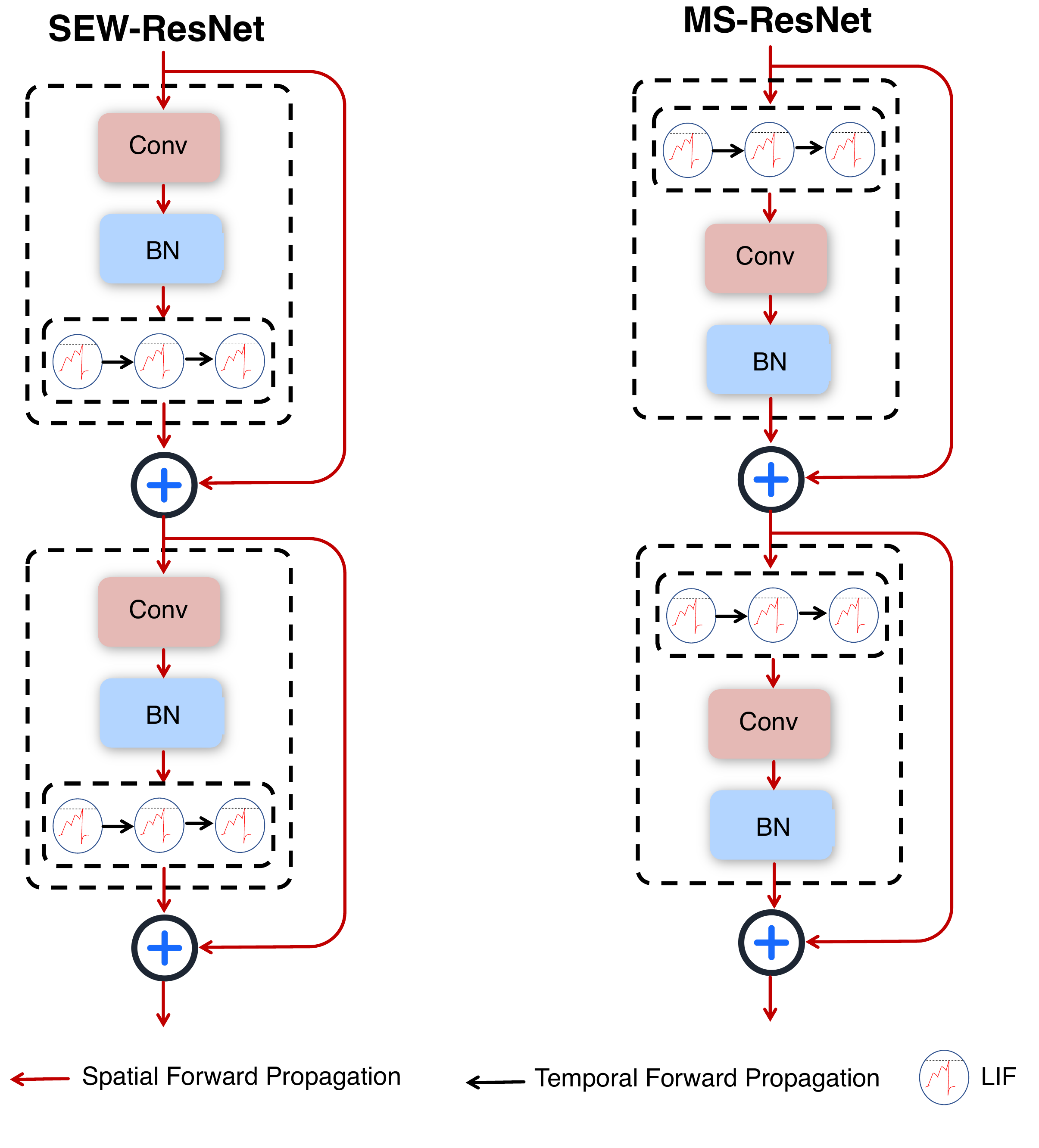}
    \caption{MS-ResNet and SEW-ResNet.}
    \label{fig:resnet}
\end{figure}

To address this issue, SEW-ResNet and MS-ResNet establish shortcut connections between spikes or membrane potentials of different layers, respectively. SEW-ResNet builds shortcuts between the output spikes of different layers, thereby obtaining an identity mapping. On the other hand, MS-ResNet constructs shortcuts between the membrane potentials of spiking neurons in different layers. Fig.~\ref{fig:resnet} illustrates the differences between these two methods. By building residual learning on membrane potentials, MS-ResNet achieves higher accuracy than SEW-ResNet. 

\section{Training}
\subsection{Pretraining on Spiking Token Mixer (STM)}
\label{appendix:pretrain}
In STM, the architecture is similar to Spiking Transformers, consisting of a Spiking Patch Embedding (SPS) layer followed by a Transformer-like architecture. However, unlike Spiking Transformers, the SPS layer in STM is not followed by a self-attention layer. We pre-train the STM on ImageNet-1K for 300 epochs, aligning with other SNNs that are pre-trained on ImageNet. The input size is set to 224 × 224, and the batch size is set to 128 or 256 during the 310 training epochs with a cosine-decay learning rate whose initial value is 0.0005. The optimizer used is Lamb. The SPS module divides the image into N = 196 patches. Standard data augmentation techniques, such as random augmentation and mixup, are also employed during training. We compare our method with other SNN methods of similar size trained on ImageNet-1K, as shown in Tab.~\ref{table_imagenet_result}.

\begin{table}[ht]
\caption{Performance on ImageNet-1K \cite{deng2009imagenet}. Note, ``Spike", ``Para", and ``Step" in all Table headers of this paper denote ``Spike-driven", ``Parameters", and ``Timestep". }
\begin{center}
\begin{tabular}{cccccc}
\toprule
\multicolumn{1}{c}{ Methods} &\multicolumn{1}{c}{  Architecture}
&\multicolumn{1}{c}{  \makecell{Spike}}
&\multicolumn{1}{c}{  \makecell{Param\\ (M)}}
&\multicolumn{1}{c}{ \makecell{Step}}
&\multicolumn{1}{c}{  Acc.(\%)}\\
\midrule
\multirow{2}{*}{ANN2SNN}   & ResNet-34 \cite{rathi2020Enabling} & \cmark & 21.8 & 250 & 61.5 \\
  & VGG-16 \cite{wu2021progressive} & \cmark & - & 16 & 65.1\\ \hline

     \multirow{3}{*}{SCNN} & SEW-Res-SNN~\cite{fang2021deep} &\xmark  & 25.6 & 4 & 67.8 \\

   & MS-Res-SNN~\cite{hu2024advancing} & \cmark &21.8 & 4 & 69.4 \\ 

    & Att-MS-Res-SNN~\cite{yao2023attention}  & \xmark &22.1 & 1 & 69.2 \\
    
\midrule
Spiking-&\multicolumn{1}{c}{\multirow{1}{*}{Spikformer~\cite{zhou2022spikformer}}} &\xmark & 16.8 & 4 & 70.2\\

Transformer&\multicolumn{1}{c}{\multirow{1}{*}{Spike-driven Transformer~\cite{yao2024spike}}}&{\cmark}&{16.8}& 4 & 72.3\\ \midrule

SMLP &\multicolumn{1}{c}{\multirow{1}{*}{\textbf{STM(Ours)}}}&{\cmark}&{12.2}& 4 & 72.1\\ \bottomrule

\hline
\end{tabular}
\end{center}
\label{table_imagenet_result}
\end{table}

\subsection{Stage-wise Training of the End-to-end Model}
\label{appendix:stage_training}
Although our model is end-to-end differentiable, directly implementing end-to-end training can lead to unstable loss convergence. To mitigate this issue, we divide our training process into three stages:
\paragraph{Stage 1: Perception Module Training}
In the first stage, we focus on training the perception module while disabling the prediction and planning modules. This means that the prediction module within the network remains inactive, and the decoder does not output instance flow information related to prediction. During this stage, we use a learning rate of $1e-3$ and train the model for 20 epochs.

\paragraph{Stage 2: Prediction Module Training}
In the second stage, we enable the prediction module, which predicts the movement of instances on the BEV for the next three steps. We train the model for an additional 10 epochs during this stage.
\paragraph{Stage 3: Planning Module Training}
Finally, we start training the planning module, using real trajectories to supervise the model's planning results. In both the prediction and planning stages, we set the learning rate to $2e-4$ to fine-tune the model from the perception stage.
Throughout all stages, we employ the Adam optimizer to update the model parameters.
By incrementally introducing the prediction and planning modules, we ensure that each component is well-trained before integrating them into the end-to-end framework. 
\section{Experiments}

\subsection{Datasets}
\label{appendix:dataset}
The nuScenes dataset~\cite{caesar2020nuscenes} is a comprehensive, publicly available dataset tailored for autonomous driving research. It encompasses 1,000 manually selected driving scenes from Boston and Singapore—cities renowned for their dense traffic and complex driving environments. Each scene, lasting 20 seconds, showcases a variety of driving maneuvers, traffic conditions, and unforeseen events, reflecting the intricate dynamics of urban driving. The dataset aims to foster the development of advanced methods that ensure safety in densely populated urban settings, featuring dozens of objects per scene. It includes around 1.4 million camera images, 390,000 LIDAR sweeps, 1.4 million RADAR sweeps, and 1.4 million object-bounding boxes across 40,000 keyframes.

\noindent\textbf{Data Preprocessing.} In preparation for modeling, we preprocess the images by cropping and resizing the original dimensions from $\mathbb{R}^{3 \times 900 \times 1600}$ to $\mathbb{R}^{3 \times 224 \times 480}$. Additionally, we incorporate temporal data by including the past three frames for each sequence, denoted as $I_t^n \in \mathbb{R}^{3 \times 224 \times 480}$, where $t \in\{1,2,3\}$ represents the frame indices, and $n \in\{1, \ldots, 6\}$ indexes the cameras.

\subsection{Evaluation Metrics}
\label{appendix:metrics}
In this section, we introduce the metrics used to assess the predictive capabilities of our model for video prediction. Specifically, we utilize Intersection over Union (IoU), Panoptic Quality (PQ), Recognition Quality (RQ), and Segmentation Quality (SQ) as defined by Kim et al.~\cite{kim2020video}.

\paragraph{Intersection over Union (IoU)}
The IoU metric evaluates the overlap between the predicted and ground truth segmentation masks. It is calculated as:
\begin{equation}
IoU = \frac{|\text{Prediction} \cap \text{Ground Truth}|}{|\text{Prediction} \cup \text{Ground Truth}|}
\end{equation}
where \( |\cdot| \) denotes the number of pixels in the respective set.

\paragraph{Panoptic Quality (PQ)}
Panoptic Quality (PQ) is a comprehensive metric that combines both segmentation and recognition quality. It accounts for both true positive matches and penalizes false positives and false negatives. PQ is defined as:
\begin{equation*}
PQ = \frac{\sum_{(p,g) \in TP} IoU(p, g)}{|TP| + \frac{1}{2}|FP| + \frac{1}{2}|FN|}
\end{equation*}
where \( TP \) represents the set of true positive matches between predicted segments \( p \) and ground truth segments \( g \), \( FP \) denotes false positives, and \( FN \) denotes false negatives.

\paragraph{Recognition Quality (RQ)}
Recognition Quality (RQ) measures the accuracy of object recognition and classification. It is given by:
\begin{equation*}
RQ = \frac{|TP|}{|TP| + \frac{1}{2}|FP| + \frac{1}{2}|FN|}
\end{equation*}

\paragraph{Segmentation Quality (SQ)}
Segmentation Quality (SQ) evaluates the quality of the predicted segments' spatial accuracy. It is defined as:
\begin{equation*}
SQ = \frac{\sum_{(p,g) \in TP} IoU(p, g)}{|TP|}
\end{equation*}
\subsection{More Visualizations}
\label{appendix:visual}
\begin{figure}
    \centering
    \includegraphics[width=\linewidth]{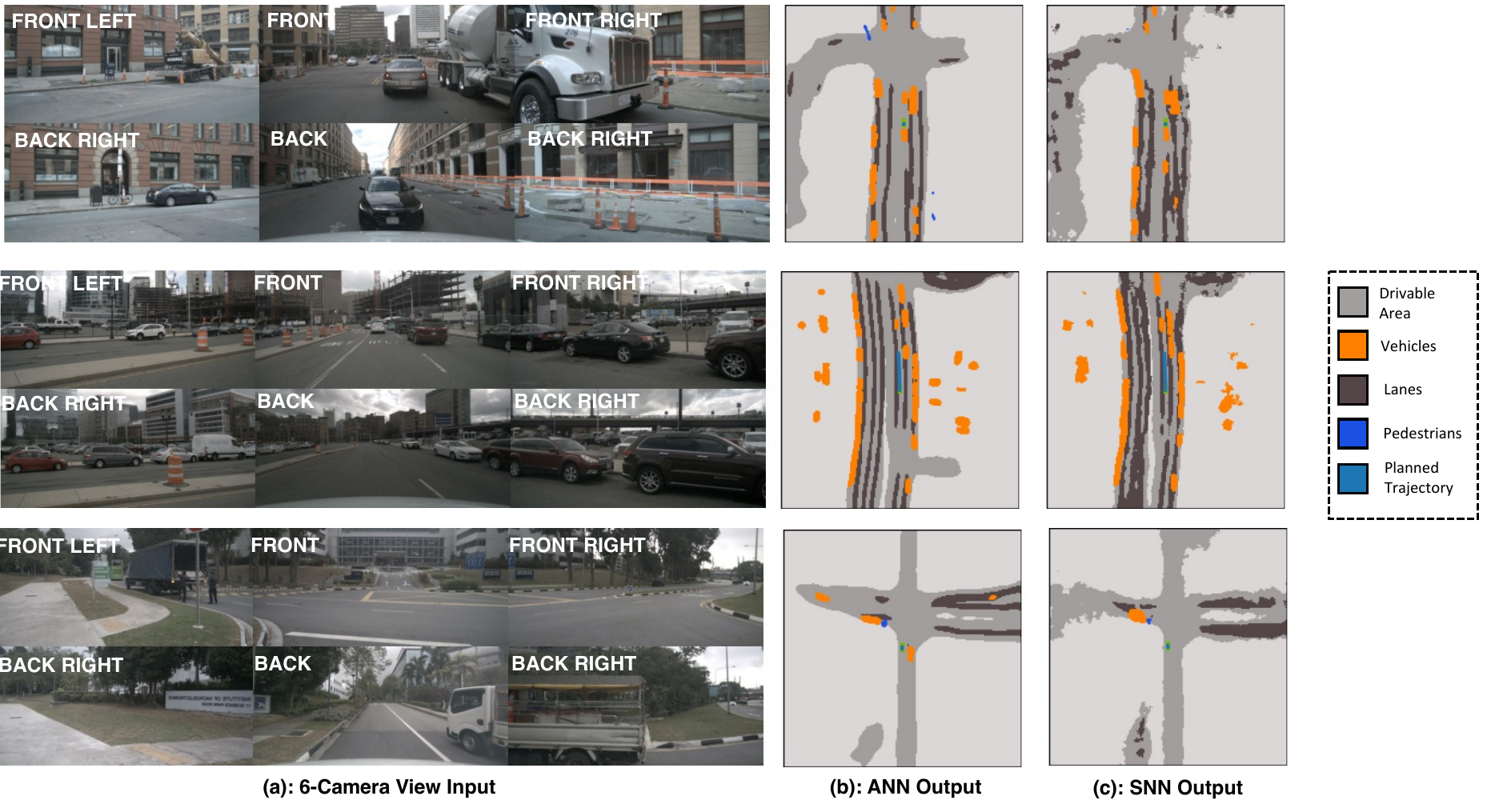}
    \caption{\textbf{More Qualitative Results of the SAD Model on the nuScenes Dataset.} (a) displays six camera view inputs utilized by the model. (b) illustrates the planning result of the ANN model, and (c) presents the planning results of our SAD model. The comparison shows that our SAD model can achieve performance comparable to that of the ANN model and successfully generate a safe trajectory.}
    \label{fig:visual2}
\end{figure}

We present three additional visual examples in Fig~\ref{fig:visual2} to further showcase the performance of our SAD model on the nuScenes dataset. These examples demonstrate the model's ability to generate safe and collision-free trajectories in various driving scenarios, highlighting its effectiveness in autonomous driving applications.

\section{Detail of Cost Function}
\label{appendix:cost}
In the method section, we briefly introduced the Cost Function $f$, a multi-component function designed to evaluate the suitability of planned trajectories for self-driving vehicles. We follow the cost function in ST-P3~\cite{hu2022st}, each component of the cost function addresses a specific aspect of the trajectory planning, as detailed below:

\textbf{Safety Cost.} This component ensures that the planned trajectories do not result in collisions. The model must account for the dynamic positioning of other vehicles and road elements, avoiding overlap with the grids they occupy. Furthermore, a safe distance must be maintained, especially at high velocities, to prevent any potential accidents.

\textbf{Cost Volume.} Road environments are complex, with numerous unpredictable elements affecting trajectory planning. It is impractical to manually enumerate all possible scenarios and associated costs. To address this, we utilize a learned cost volume produced by the prediction module head mentioned earlier. To prevent the cost volume from disproportionately influencing the trajectory evaluation, we clip its values to maintain balanced decision-making.

\textbf{Comfort and Progress.} To ensure that the trajectories are not only safe but also comfortable for passengers, this component penalizes excessive lateral acceleration, jerk, and high curvature. Additionally, the efficiency of the trajectory is critical; thus, paths that effectively progress towards the target destination receive positive reinforcement.

These components collectively ensure that the cost function comprehensively assesses both the safety and quality of the trajectories planned by the self-driving vehicle.

\section{Theoretical Energy Consumption Calculation}
\label{appendix:energy}

In the experimental section of our paper, we utilize energy consumption as a metric to evaluate the efficiency of various models. This appendix outlines the methodology employed to compute the theoretical energy consumption of our SNN architecture. The calculation process involves two main steps: first, determining the synaptic operations (SOPs) for each architectural component, and second, estimating the cumulative energy consumption based on these operations.

The synaptic operations within each block of the SNN are calculated using the following equation:
\begin{equation}
  \operatorname{SOPs}(l) = fr \times T \times \operatorname{FLOPs}(l)
\end{equation}
where \(l\) denotes the ordinal number of the block within the SNN, \(fr\) represents the firing rate of the input spike train to the block, \(T\) indicates the time step of the neuron, and \(\operatorname{FLOPs}(l)\) refers to the floating-point operations within the block, primarily consisting of multiply-and-accumulate (MAC) operations.

To compute the SNN's theoretical energy consumption, we consider both MAC and spike-based accumulation (AC) operations, utilizing 45 nm technology. The energy costs are \(E_{MAC} = 4.6 \, \text{pJ}\) for MAC operations and \(E_{AC} = 0.9 \, \text{pJ}\) for AC operations. Based on the methodologies outlined in \cite{yao2023attention}, we calculate the energy consumption of the SNN as follows:
\begin{equation}
 \begin{aligned}
E_{SNN} &= E_{MAC} \times \mathrm{FLOP}_{\mathrm{SNN}_\mathrm{Conv}}^1 \\
&+ E_{AC} \times \left(\sum_{n=2}^N \mathrm{SOP}_{\mathrm{SNN}_\mathrm{Conv}}^n + \sum_{m=1}^M \mathrm{SOP}_{\mathrm{SNN}_\mathrm{FC}}^m\right)
\end{aligned}
\end{equation}
Here, \(N\) and \(M\) denote the number of convolutional (Conv) and fully connected (FC) layers, respectively. The terms \(\mathrm{FLOP}_{\mathrm{SNN}_\mathrm{Conv}}\) and \(\mathrm{SOP}_{\mathrm{SNN}_\mathrm{Conv}}\), \(\mathrm{SOP}_{\mathrm{SNN}_\mathrm{FC}}\) represent the FLOPs for the first Conv layer and SOPs for the subsequent \(n^{th}\) Conv and \(m^{th}\) FC layers.

For the ANN model, the formula for computing the theoretical energy consumption is more straightforward, as all neurons operate in floating point:
\begin{equation}
E_{ANN} = E_{MAC} \times \mathrm{FLOP}_{\mathrm{ANN}}
\end{equation}
Here, $\mathrm{FLOP}_{\mathrm{ANN}}$ represents the FLOPs of the ANN model.

\end{document}